\documentclass[conference]{IEEEtran}
\IEEEoverridecommandlockouts
\usepackage{cite}
\usepackage{amsmath,amssymb,amsfonts}
\usepackage{algorithmic}
\usepackage{graphicx}
\usepackage{textcomp}
\usepackage{xcolor}

\usepackage{pifont}        % dings
\usepackage{makecell}      % cells in tables
\usepackage{multirow}      % for vertical centering
\usepackage{subcaption}    % for multi-column figures
\usepackage[table]{xcolor}
\usepackage{booktabs}
\usepackage{array}
\usepackage{tikz}
\usepackage[ruled,linesnumbered]{algorithm2e}

\def\BibTeX{{\rm B\kern-.05em{\sc i\kern-.025em b}\kern-.08em
    T\kern-.1667em\lower.7ex\hbox{E}\kern-.125emX}}

\begin{document}

\title{Have I Seen You? Embedding Behavior Signals Synthetic Face Dataset Membership}

% \author{
% \IEEEauthorblockN{Anonymous Submission}
% }

% \author{
% \IEEEauthorblockN{\phantom{aaaaaaa}Paweł Borsukiewicz \phantom{aaaaaaa}}
% \IEEEauthorblockA{\textit{SnT} \\
% \textit{University of Luxembourg}\\
% Luxembourg, Luxembourg \\
% pawel.borsukiewicz@uni.lu}
% \and
% \IEEEauthorblockN{\phantom{aaaaaaa}Daniele Lunghi\phantom{aaaaaaa}}
% \IEEEauthorblockA{\textit{SnT} \\
% \textit{University of Luxembourg}\\
% Luxembourg, Luxembourg \\
% daniele.lunghi@uni.lu}
% \and
% \IEEEauthorblockN{\phantom{aaaaaaa}Wendkûuni C. Ouédraogo\phantom{aaaaaaa}}
% \IEEEauthorblockA{\textit{SnT} \\
% \textit{University of Luxembourg}\\
% Luxembourg, Luxembourg \\
% wendkuuni.ouedraogo@uni.lu}
% \and
% \IEEEauthorblockN{\phantom{aaaaaaaaaaa}Jacques Klein\phantom{aaaaaaa}}
% \IEEEauthorblockA{\phantom{aaaaa}\textit{SnT} \\
% \textit{\phantom{aaaaa}University of Luxembourg}\\
% \phantom{aaaaaa}Luxembourg, Luxembourg \\
% \phantom{aaaaa}jacques.klein@uni.lu}
% \and
% \IEEEauthorblockN{Tegawend\'e F. Bissyand\'e}
% \IEEEauthorblockA{\textit{SnT} \\
% \textit{University of Luxembourg}\\
% Luxembourg, Luxembourg \\
% tegawende.bissyande@uni.lu}
% \and
% \IEEEauthorblockN{\phantom{Author Name}}
% \IEEEauthorblockA{\phantom{\textit{SnT}} \\
% \phantom{\textit{University of Luxembourg}}\\
% \phantom{Luxembourg, Luxembourg} \\
% \phantom{email@uni.lu}}
% }

\author{
    Paweł Borsukiewicz, Daniele Lunghi, Wendkûuni C. Ouédraogo,
    Jacques Klein, Tegawendé F. Bissyandé \\
    %\thanks{Corresponding author}
    \normalsize University of Luxembourg, Luxembourg\\
    {\tt\scriptsize \{pawel.borsukiewicz, daniele.lunghi, wendkuuni.ouedraogo, jacques.klein, tegawende.bissyande\}@uni.lu}
    \vspace{-1em}
}

\maketitle

\begin{abstract}

Synthetic face datasets are increasingly used to reduce privacy exposure and data access constraints in biometric recognition. Yet the generators that produce these datasets are trained on real faces, so synthetic data may still reveal their real source data. We study this risk through a dataset-level membership inference attack that first identifies the synthetic dataset used to train a face recognizer and then infers the real dataset used to train the generator. Across 11 face recognition models, 11 synthetic datasets, and 7 real datasets, the attack recovers the synthetic training dataset in 100\% of cases and identifies the generator's source dataset in 54.5\% of cases. These results show that synthetic data can retain dataset-level traces of real training data and that privacy-preserving deployment requires stronger leakage mitigation.
\end{abstract}


\section{Introduction \& Background}

%As underscored in Article 12 of the Universal Declaration of Human Rights, privacy is a fundamental right.
%Nowadays, with the rise in the use of biometric systems, this freedom from unlawful interference is increasingly being challenged. While many can claim the tangible benefits of automated recognition systems for law enforcement and their practical applications as access control mechanisms for our smartphones, the full scope of data processing and surveillance remains unknown. 

Face recognition has become ubiquitous in security and consumer applications, yet it relies on large biometric datasets whose collection and sharing raise persistent privacy and compliance concerns. Lately, synthetic face datasets have emerged as an attractive alternative, reducing direct exposure of real identities and easing data access constraints. Recent work shows that synthetic training data can support recognition performance comparable to, and in some cases better than, models trained on real data~\cite{borsukiewicz2026beyond}. That progress makes synthetic data a practical option for biometric model development.
%, but it does not eliminate privacy risk.

%The remaining risk stems from how synthetic datasets are created.

However, the privacy benefit is incomplete. Both GAN- and diffusion-based generators are trained on real facial images, which means that traces of the source data may survive in the generated distribution. Membership inference attacks (MIAs) exploit such traces by asking whether specific data contributed to model training~\cite{shokri2017membership}. In the biometric setting, successful inference could reveal sensitive information about individuals or about the hidden provenance of a deployed model.

Prior work on synthetic face leakage studied identity-level exposure by retrieving real samples that are unusually similar to synthetic ones~\cite{shahrezaunveiling}. That setting is important, but it assumes that the attacker already knows which synthetic dataset trained the recognizer, and the leakage signal itself is concentrated in a small fraction of images~\cite{borsukiewicz2026beyond}. 
In this paper, we address the prior question, namely, which synthetic datasets were used to train the model, and then extend the attack one step further upstream to the real dataset used to train the generator. Across eleven synthetic datasets, 7 real datasets, and 11 face recognition models, our attack recovers the synthetic training dataset in every case and reveals the generator's source dataset in 54.5\% of cases.
Combined with~\cite{shahrezaunveiling}, this facilitates an end-to-end pipeline from a deployed recognizer to real identities.

\begin{figure*}[h]
    \centering
    \includegraphics[width=.85\linewidth]{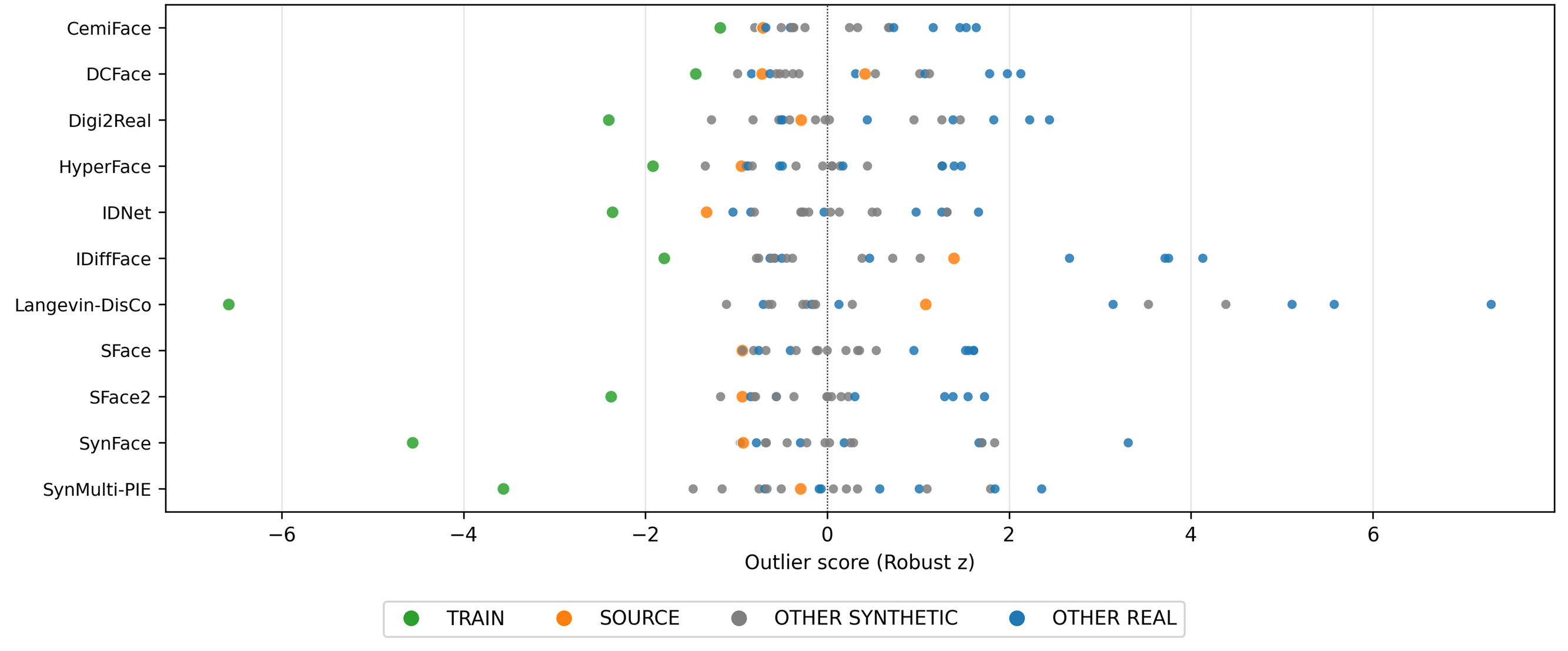} 
    \caption{Outlier score per dataset. Synthetic data used in training (TRAIN) is indicated on the y-axis.}
    \label{fig:results1}
    \vspace{-1em}
\end{figure*} 

\vspace{-0.25em}
\begin{algorithm}
\caption{Membership Inference Attack}
\label{alg:alg1}
\KwIn{Trained model $f$, candidate datasets $\mathcal{D} = \{D_1, \dots, D_K\}$, $N = 500$}
\KwOut{Robust z-score $\tilde{z}_i$ ranking for each dataset $D_i$}

Step 1: Compute mean probe-to-centroid similarity for each dataset\;
\For{each $D_k \in \mathcal{D}$}{
    Sample $N$ random gallery identities $\mathcal{I}_k$ and $N$ random probe identities $\mathcal{P}_k$ (disjoint)\;
    Compute L2-normalized gallery centroid $\mathbf{c}_k$: $\mathbf{c}_k \gets \frac{1}{N}\sum_{i \in \mathcal{I}_k} f(i)$, \quad $\mathbf{c}_k \gets \mathbf{c}_k / \|\mathbf{c}_k\|$\;
    Find mean probe-to-centroid cosine similarity $x_k$: $x_k \gets \frac{1}{N} \sum_{j \in \mathcal{P}_k} \dfrac{f(j) \cdot \mathbf{c}_k}{\|f(j)\|}$
}

Step 2: Compute robust z-score for target dataset $x_t$\;
\For{each $D_t \in \mathcal{D}$}{
    $\tilde{x} \gets \mathrm{median}(\{x_i\}_{i \neq t})$ \;
    $\mathrm{MAD} \gets \mathrm{median}(|x_i - \tilde{x}|)_{i \neq t}$\;
    $\tilde{z}_t \gets \dfrac{0.6745*(x_t - \tilde{x})}{\mathrm{MAD}}$\;
}

Step 3: Rank z-scores; 

\end{algorithm}
\vspace{-1.25em}

\section{Methodology}

We aim to connect a deployed recognition model to its data on two levels: (i) identify which synthetic dataset was used to train the model, and (ii) given a synthetic-trained model, identify which real dataset was used to train the generator that produced the synthetic data. 
To do so, we propose a simple attack (Algorithm~\ref{alg:alg1}), based on one assumption and one hypothesis. First, in line with previous works on loss function optimization~\cite{wang2018cosface}, we assume that each model should yield the best separation of similarity scores between non-mated images (of distinct identities) for its training data, thereby resulting in the lowest robust z-score values, defined as the difference between the target dataset's median score and the median of the remaining datasets' scores, divided by the Median Absolute Deviation (MAD) of the remaining datasets' scores. Second, we hypothesize that because synthetic data inherits the distribution of the generator's training data, this property may carry over to the real source data, producing a secondary leakage signal.
To make the attack realistic, we assume a black-box adversary with query access to the model's embeddings.

%We expect that, due to the inherent similarity between synthetic data and the generator's training data, this property could be inherited by the real data, resulting in leakage of the real dataset. 

%We propose a simple yet effective dataset-level MIA against synthetic facial recognition datasets (Algorithm~\ref{alg:alg1}). Our goal is to determine which dataset was used to train a facial recognition model. Additionally, if we learn that a model was trained on synthetic data in the first stage, we would like to know which real dataset was included in the generator's training set. 

For our attack, we have selected the iResNet50 backbone and trained eleven face recognition models with synthetic datasets (Table~\ref{tab:datasets_used}) as their training sets, one dataset per model. 
%We selected CosFace loss function with a margin of $m=0.35$ and trained for 36 epochs with early stopping. We have also initially tested ArcFace, however, we abandoned it due to training instability. 
For each face recognizer, we have computed embeddings across all synthetic and seven real datasets. We have selected FFHQ and CASIA-WebFace due to their common use in training synthetic dataset generators, and LFW, CPLFW, CALFW, CFP-FP, and AgeDB-30 to enable a baseline comparison on unseen data. 

\begin{table}[h]
\centering
\footnotesize
\caption{Synthetic datasets and their deep generative models' training datasets.}
\setlength{\tabcolsep}{3pt}
\label{tab:datasets_used}
\begin{tabular}{lc}
\toprule
Synthetic dataset & Generator training data (Source dataset) \\
\midrule
\makecell[l]{SynFace, IDiff-Face, \\ SynMulti-PIE, Langevin-Disco}   & FFHQ \\
IDNet, CemiFace, SFace2, Sface             & CASIA-WebFace \\
DCFace                               & FFHQ + CASIA-WebFace \\
HyperFace                            & FFHQ + CelebA + WebFace4M \\
Digi2Real                            & FFHQ + CelebA + WebFace42M \\
% SynFace~\cite{qiu2021synface}    & FFHQ \\
% IDiff-Face~\cite{boutros2023idiff}    & FFHQ \\
% SynMulti-PIE~\cite{colbois2021synmultipie} & FFHQ \\
% IDNet~\cite{kolf2023identity}      & CASIA-WebFace \\
% CemiFace~\cite{sun2024cemiface}   & CASIA-WebFace \\
% SFace2~\cite{boutros2024sface2}     & CASIA-WebFace \\
% SFace~\cite{boutros2022sface}      & CASIA-WebFace \\
% DCFace~\cite{kim2023dcface}     & FFHQ + CASIA-WebFace \\
% HyperFace~\cite{shahreza2024hyperface}    & FFHQ + CelebA + WebFace4M   \\
% Digi2Real~\cite{george2025digi2real}    & FFHQ + CelebA + WebFace42M \\
\bottomrule
\end{tabular}
\vspace{-1em}
\end{table}

% \begin{equation}
%     z = \frac{0.6745*(x - \tilde{x})}{\text{median}\left(|x_i - \tilde{x}|\right)},
% \end{equation}
\section{Results \& Discussion}

\noindent\textbf{Synthetic training datasets: }Our results (Figure~\ref{fig:results1}) show that the synthetic training datasets consistently achieve by far the lowest robust z-score values, clearly revealing the training data. Consequently, any potential attacker who observed a value less than -2 in our setting would be virtually certain that a given model was trained on particular data. As our algorithm is compatible with a black-box scenario, where API access exposes the sample's embedding, one could use this information to retrain their own model and leverage white-box access for any further analysis. 

%Negative values correspond to data points below the median, indicating lower average similarity scores between non-mated (of distinct identities) samples. 
%This phenomenon stems from a training procedure in which the models learn to separate the training set samples and are consequently most efficient on their respective training data.

%\input{Figures/f2}

\noindent\textbf{Real source datasets: }We evaluated whether the fact that the synthetic data follows the distribution of its underlying real dataset is sufficient to deduce its use during generator training. For that matter, we could see (Figure~\ref{fig:results1}) that the source datasets are, on average, achieving lower scores than other real datasets. Analyzing the lowest scores across the real datasets has revealed a training set 6 out of 11 times. Some of the failures could be attributed to the multi-dataset training set, e.g., Digi2Real, where the overlap of samples among CelebA, WebFace42M, and CASIA-WebFace could have resulted in CASIA-WebFace receiving a lower score than the evaluated training set, FFHQ (-0.41 vs. -0.29). Ultimately, our attack demonstrates that real data is at risk and that training the model on synthetic data requires additional safeguards to prevent leakage of the real dataset.

%\noindent\textbf{Stronger separability of synthetic data: }
%Synthetic data tends to achieve a lower robust z-score, suggesting higher structural similarity to the training data

\section{Conclusion \& Future Work}

This paper highlights the privacy risks posed by attacks against synthetic facial recognition datasets. Using the novel MIA algorithm, we showcase that a potential adversary can easily identify the synthetic dataset used in training and has a relatively high chance of deducing the generator's real data.

Future work spans two dimensions. Firstly, we will search for suitable measures to prevent identity and dataset leakage. Secondly, to better assess protective techniques, more sophisticated algorithms could be explored to address currently limited efficiency against the underlying real datasets.

\section*{Acknowledgments}
\begin{itemize}
    \item This research was funded by the Luxembourg Army.
    \item The authors acknowledge the use of AI-based tools to correct typographical errors and improve clarity.
\end{itemize}

\bibliographystyle{IEEEtran}
\bibliography{main}

\end{document}